\documentclass{article}
\usepackage{ijcai21}

\usepackage{times}
\usepackage{soul}
\usepackage{url}
\usepackage{microtype}
\usepackage[hidelinks]{hyperref}
\usepackage[utf8]{inputenc}
\usepackage[small]{caption}
\usepackage{subcaption}
\usepackage{graphicx}
\usepackage{booktabs}
\usepackage{algorithm}
\usepackage{algorithmic}
\usepackage{courier}
\usepackage{color}
\usepackage{amsmath,amsfonts,amssymb,amsthm,amsopn}
\usepackage{epsfig}
\usepackage{booktabs}
\usepackage{array}
\usepackage{multicol}
\usepackage{threeparttable}
\usepackage{epstopdf}
\usepackage{listings}
\usepackage{multirow}
\usepackage{ragged2e}
\usepackage{xassoccnt}

\newtheorem{example}{Example}
\newcommand{\secref}[1]{Section \ref{#1}}
\newcommand{\figref}[1]{Figure \ref{#1}}
\newcommand{\eqnref}[1]{Eq. (\ref{#1})}
\newcommand{\tabref}[1]{Table \ref{#1}}
\newcommand{\exref}[1]{Example \ref{#1}}
\newcommand{\cut}[1]{}
\newcommand{\citealp}[1]{\citeauthor{#1}~\shortcite{#1}}

\newcommand\BibTeX{B\textsc{ib}\TeX}
\newcommand{\sgn}{\text{sgn}}

\def\BibTeX{{\rm B\kern-.05em{\sc i\kern-.025em b}\kern-.08em
    T\kern-.1667em\lower.7ex\hbox{E}\kern-.125emX}}

\title{Statistically Profiling Biases in Natural Language Reasoning Datasets and Models}

\author{
Shanshan Huang
\and
Kenny Q. Zhu
\affiliations 
Shanghai Jiao Tong University \\
Shanghai Jiao Tong University
\emails
huangss\_33@sjtu.edu.cn \\
kzhu@cs.sjtu.edu.cn
}

\begin{document}
\maketitle
\begin{abstract}
Recent work has indicated that many natural language understanding and 
reasoning datasets contain statistical cues that
may be taken advantaged of by NLP models whose
capability may thus be grossly overestimated. 
To discover the potential weakness in the models, some human-designed 
stress tests have been proposed but they are expensive to create 
and do not generalize to arbitrary models. 
We propose a light-weight and general statistical profiling framework, 
ICQ (I-See-Cue), which automatically identifies possible biases
in any multiple-choice NLU datasets without 
the need to create any additional test cases, and further evaluates
through blackbox testing the extent to which models may exploit these biases. 
\end{abstract}

\section{Introduction}
\label{sec:intro}
Deep neural models have shown to be effective in 
a large variety of natural language inference~(NLI)
tasks~\cite{bowman2015large,wang2018glue,mostafazadeh2016corpus,roemmele2011choice,zellers2018swag}. Many of these tasks
are discriminative by nature, such as predicting a class label or
an outcome given a textual context, as shown in the following example:

\begin{example}\label{exp:snli}
Natural language inference in SNLI dataset, with ground truth bolded.
\begin{description}
\item{Premise:} A swimmer playing in the surf watches a low flying airplane headed inland. 
\item{Hypothesis:} Someone is swimming in the sea.
\item{Label:} \textbf{a) Entailment.} b) Contradiction.  c) Neutral.
\end{description}
\end{example}

The number of candidate labels may vary. Humans solve such questions by
reasoning the logical connections between the premise and the hypothesis,
but previous work~\cite{naik2018stress,schuster2019towards} 
has found that some NLP models can solve the questions
fairly well by looking only at the hypothesis (or ``conclusion'' in some work)
in the datasets.
It is widely speculated that this is because in many datasets, 
the hypotheses are manually crafted and may contain artifacts that
would be predictive of the correct answer. 
Such ``hypothesis-only'' tests can identify problematic questions
in the dataset if the question can be answered correctly without 
the premise. While such a method to evaluate the quality of
a dataset is theoretically sound, 
it i) usually relies on training a heavy-weight model such as Bert, which
is costly to evaluate, ii) does not provide explanation why the question is 
a culprit, and iii) cannot be used to evaluate a model since a model that
can make a correct prediction using only the hypothesis is not necessarily a
bad model: it is just not given the complete data.


Inspired by black-box testings in software engineering, 
CheckList~\cite{checklist2020acl} assesses the weakness of 
models without the need to know the details of the model. It does so by
providing additional stress test cases according to predefined 
linguistic features. Unfortunately, to ensure the correctness of
these additional cases, the templates must be carefully crafted
with substantial restrictions, thus limiting the testing space and
complicating the implementation. 
Furthermore, with CheckList, you only get to know what the model 
is incapable of doing but won't know what the model has 
learned from the data.

\begin{figure}[th]
\centering
\includegraphics[width=0.6\columnwidth]{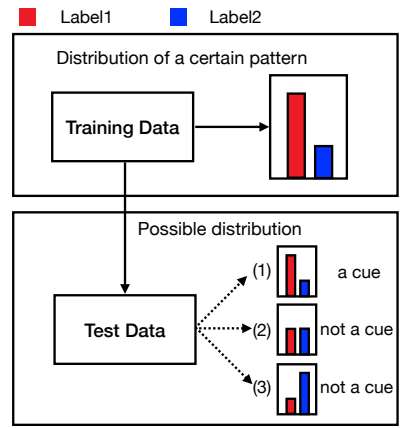}
\caption{Example of a {\em cue}. }
\label{fig:cue_def}
\end{figure}

In this paper, our view is that the existing test sets for these
tasks are not sufficiently exploited. Why do we go the extra mile to
generate new test cases which are potentially incorrect, when we can
test the models using existing test sets but from different perspectives?
With this objective in mind, we propose ICQ 
(``I-see-cue''), an open-source evaluation 
framework~\footnote{\url{http://anonymized.for.blind.review}} for 
evaluating both the dataset and the corresponding models. In this framework, 
one test dataset can be seen as multiple test sets from different perspectives. 

Previous studies~\cite{gururangan2018annotation,sanchez2018behavior,poliak2018hypothesis} 
showed that statistical bias on linguistic features 
(e.g., sentiment, repetitive words and even shallow n-grams)
which have statistical correlation with specific labels 
in benchmark datasets can be predictive of the correct 
answer. We call such biases artificial spurious \textit{cues} when
they appear both in the training and test datasets with a similar distribution
over the prediction values.
We illustrate this in~\figref{fig:cue_def}. 
Once these cues are neutralized from the test data, 
previously successful models may degrade substantially
on it, suggesting that the model has taken advantage of 
the biased feature and is hence not as robust as assumed 
against with such cues.

In summary, this paper makes the following contributions:
\begin{itemize}
\item we provide a light-weight but effective method to uncover
the statistical biases and cues in NL reasoning datasets;

\item we propose two simple tests to quantitatively and visually
assess whether a given model has taken advantage of a
spurious cue when making predictions;

\item we evaluate the statisical bias issues comprehensively on
10 popular NLR datasets and 4 models and not only reaffirm
some of the findings from previous work but also discover some
new perspectives in these datasets and models;

\item we created an online demonstration system to showcase
the results and invite users to evaluate their own datasets and models.
\end{itemize}


\section{Preliminary Definition}
\label{sec:formulation}
We define an instance $x$ of a natural language reasoning (NLR) task 
dataset $X$ as
\begin{equation}
    x = (p, h, l) \in X, \label{eq:nli}
\end{equation}
\noindent
where $p$ is the context against which to do the reasoning ($p$ corresponds 
to ``premise'' in~\exref{exp:snli});
$h$ is the hypothesis given the context $p$; 
$l \in \mathcal{L}$ is the label that 
depicts the type of relation between $p$ and $h$. 
The size of the relation set $\mathcal{L}$ varies with tasks. 


There is another type of natural language reasoning tasks which 
are also in the form of multiple-choice questions, 
but their choices are a fixed set of labels, as shown below. 
\begin{example}\label{exp:roc}
A story in ROCStory dataset, with ground truth bolded~\cite{mostafazadeh2016corpus}.
\begin{description}
\item{Context:} Rick grew up in a troubled household. 
He never found good support in family, and turned to gangs.           
It was n't long before Rick got shot in a robbery.             
The incident caused him to turn a new leaf.
\item{Ending 1:} He joined a gang. 
\item{Ending 2:}  \textbf{He is happy now.}
\end{description}
\end{example}

We can transform the this case into two separate 
problem instances, still
in the same form as in \eqnref{eq:nli}, 
$u_1=(context, ending1, false)$ and $u_2=(context, ending2, true)$, where $L = {true, false}$.

\section{Approach}
\label{sec:approach}

\begin{figure}[th]
\centering
\includegraphics[width=0.8\columnwidth]{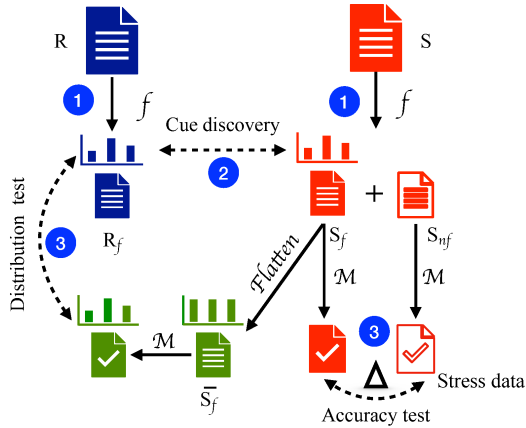}
\caption{ICQ Workflow. \textcircled{1}: data filtering phase; \textcircled{2}: cue discovery phase; 
\textcircled{3}: model probing phase.}
\label{fig:framework}
\end{figure}

The ICQ framework is illustrated in~\figref{fig:framework}.
In \textit{data filtering} phase, it extracts from the dataset
those problem instances that contain a given linguistic feature $f$. 
In \textit{cue discovery} phase, it identifies the possible cues in this dataset. 
Finally, in \textit{model probing} phase,
it does two tests: ``accuracy test'' and ``distribution test''.
Next we will discuss these phases in more details.

\subsection{Linguistic Features}
\label{sec:extract}

In this work, we consider the following linguistic features: 
Word (unigram tokens in the input sentences), 
Typos, NER (named entity recognition), 
Tense (temporal order of events), Negation, 
Sentiment, and Overlap (words occurring both in the premise and 
the hypothesis). 
The above list is by no means exhaustive, but just a starting point for users 
who can come up with additional features that are relevant
to their task or domain. 

\subsubsection{Word} 
For a dataset $X$, we collect a set of all words 
$V$ that ever exist in $X$. 
A word feature is defined as the existence of a word $w \in V$
either in the premise or the hypothesis. 
Because $V$ is generally very large, in practice, we may narrow it down
to words that sufficiently popular in $X$. That is, we may remove
words that seldom appear in $X$.
%

%

\subsubsection{Sentiment}

For each data instance $x$, we can compute its sentiment value as:
\begin{equation}
S(x) = \sgn(\sum_{w \in x} polar(w)),
\end{equation}
where $polar(w)$ is the sentiment polarity (-1, 0, or 1)
of $w$ determined by a look-up from a pretrained sentiment 
lexicon~\footnote{NLTK: \url{https://www.nltk.org}}.
We say $x$ has a positive/negative/neutral sentiment feature if $S(x)$ = 1, -1 or 0,
respectively.

\subsubsection{Tense}
We say that an instance $x$ has  
\textit{past}, \textit{present} or \textit{future} tense feature if $x$
carries one of these tenses, respectively, by the POS tag of the root verb
in $p$ or $h$. 

\subsubsection{Negation}
Previous work has observed that negative words (``no'', ``not'' or ``never'') 
may be indicative of a certain label in NLI tasks for some models.
The existence of a negation feature in $x$ is decided by dependency 
parsing~\footnote{Scipy: \url{https://www.scipy.org}}. 

\subsubsection{Overlap}
In many models, substantial word-overlap between the premise and the
hypothesis sentences causes incorrect inference, 
even if they are unrelated~\cite{mccoy2019right}. 
We define that an overlap feature exists in $x$ if there's at least one word
(except for stop words) that occurs both in $p$ and $h$. 

\subsubsection{NER}
We define the NER feature as the existence of either PER,
ORG, LOC, TIME, CARDINAL entity in $x$.
We use the NLTK ner toolkit for this purpose. 

\subsubsection{Typos}
We say an instance $x$ has typo feature if there exists at least one
typo in $x$.
We use a pretrained spelling model~\footnote{\url{https://github.com/barrust/pyspellchecker}} 
to detect all typos in a sentence. We don't distinguish the types of mispellings here. 

As we mentioned in Example \ref{exp:roc}, multiple choice questions are each split into 
two data instances with opposite labels (T or F), and the premises in these two
instances are identical. Therefore, it is not useful to detect features within the premises
alone. Consequently, for MCQ type of datasets, all the above features except for Overlap
are only applied on the hypotheses.

\subsection{Discovering Cues in Dataset}
\label{sec:evaldata}
Once we have defined the linguistic features, we can build a data filter for
each feature values. A filter takes a dataset and returns
a subset of instances associated with that feature value. For example,
there is a filter for the word ``like''; there is a filter for ``PER'' entity;
and there is filter for ``negative'' sentiment, etc. 

For each feature $f$, we apply its filter to both the training data and test data
of $X$, denoted as $R$ and $S$ in \figref{fig:framework},
resulting in $R_f$ and $S_f$. 
Only those features that appear both in the training and test data
are qualified as possible cues for a dataset.
Let $y_i$ be the number of instances with label $l_i$ in the filtered
dataset, then we can compute the bias of the label distribution for a filtered
set as the mean squared error (MSE):
\begin{equation}
mse(Y) = \frac{1}{|L|} \sum_i (y_i - \hat{y_i})^2
\end{equation}
where $\hat{y_i}$ is the mean of $y_i$. The larger $mse(Y)$, 
the more ``pointed'' the label distribution and more biased. 
Furthermore, if the filtered training set and 
the filtered test set are biased similarly,
the Jensen-Shannon Divergence~\cite{lin1991divergence} between
them is small:
\begin{equation}
jsd(R_f, S_f) = \frac{1}{2}\left (R_f\parallel M  \right )+\frac{1}{2}\left (S_f\parallel M  \right ), 
\end{equation}
where $M= \frac{1}{2}\left (R_f+S_f \right )$. 
Finally, we define a cueness score as
\begin{equation}
cue(f, X) = \frac{mse(R_f)}{\exp(jsd(R_f), S_f))}
\end{equation} 
which represents how much a dataset $X$ is biased against a feature
$f$. 

\subsection{Probing Bias in Model}
\label{sec:evalmodel}

Suppose we already know that a dataset $X$ is infected with a cue $f$ from previous
test in \secref{sec:evaldata}.
However, just because the dataset is infected with a cue doesn't mean
the model trained from this dataset necessarily exploits that cue.
Here we propose a simple method to probe any model instance trained from the
given biased dataset~\footnote{Models trained from any other 
datasets compatible in format with $X$ can also be used to probe its potential
bias on the same cue.} to see if it actually takes advantage of that cue $f$
and by how much.

We can do that through two simple tests: 
{\em accuracy test} and {\em distribution test}.
In accuracy test, we simply assess the prediction accuracies of the model
$M$ on the filtered test set and on the remaining test data, and call them
$acc(S_f)$ and $acc(S_{nf})$, respectively. The accuracy test says that if the difference
between these two accuracies, i.e, $\Delta=acc(S_f) - acc(S_{nf})$ is greater 0, then the
model is considered to be biased and to have exploited this cue. 
The value of $\Delta$ measures the extent of the bias.
 
Distribution test is a visual test. We first create a ``stress data set'' $\overline{S_f}$
by ``flattening'' the label distribution in $S_f$.  
We achieve that by replicating random instances from all labels 
except the most popular label in the filtered test
set and adding them back into the set. 
The repetition augmentation procedure stops when 
the feature distribution based on each label is balanced.
This way we have effectively removed the bias 
in the filtered test set and presumably 
posed a challenge to the model. 
Next we apply the same model on the stress test set 
to get prediction results. 
We compare the label distribution of the prediction results on 
the stress test set with the label distribution of 
the filtered training data.
The idea is, if the filtered training data contains a cue, 
its label distribution will be skewed toward a particular label.
If the model exploits this cue, it will prefer to predict
that label as much as possible, even amplifying the skewness
of the distribution, despite that the input test set has been
neutralized already. We hope to witness such an amplification
in the output distribution to capture
the weakness in the model.

The above two tests are related but not equivalent
and their outcome complement and reinforce each other. 



\section{Evaluation}
\label{sec:eval}
We first show the experimental setup and then give the results
on cue discovery as well as model probing along with some analysis.
The whole framework has been implemented into an online demo at
\url{http://anonymized.for.blind.review}.

\subsection{Setup} 
We evaluate this framework on 10 popular NLR datasets and
4 well-known models, namely FASTText (FT), ESIM (ES), BERT (BT) and RoBERTA (RB)
on these datasets. All these datasets except for SWAG and RECLOR are collected
through crowdsourcing. SWAG is generated from an LSTM-based language model.
Specifications of the datasets are listed in \tabref{tab:datasets}.

\begin{table}[th]
\scriptsize
\centering
\begin{tabular}{lcccc}\hline
Dataset & Type & Data Size & Train/Test & Human Acc\\ 
 	&	&	& Ratio	& (\%) \\ \hline
SNLI     &CLS   &  570K     & 56:1               &80.0\\
QNLI     &CLS    & 11k         &  19:1           &80.0\\
MNLI     &CLS     & 413k       &  40:1             &80.0\\
ROCStory & MCQ & 3.9k         & 1:1            &100.0  \\
COPA     &MCQ    & 1k           &  1:1         & 100.0     \\
SWAG     &MCQ   & 113k       &  4:1             & 88.0\\
RACE     & MCQ   & 100k      &  18:1              &94.5\\
RECLOR   &MCQ    &  6k          &  9:1           &63.0\\
ARCT     &MCQ    & 2k         & 3:1                &79.8\\
ARCT\_adv& MCQ & 4k         & 3:1                 & -\\
\hline
\end{tabular}
\caption{10 Datasets. Data size refers to the number of questions
in each dataset. CLS=Classification. MCQ=Multiple Choice Questions. 
By our definition, $k$-way MCQs will be split into $k$ instances 
in preprocessing.}\label{tab:datasets} 
\end{table}
 
These datasets can mainly be classified into two types of tasks. 
SNLI, QNLI, and MNLI are classification tasks, while 
ROCStory, COPA~\cite{roemmele2011choice}, SWAG~\cite{zellers2018swag}, 
RACE~\cite{lai2017race}, 
RECLOR~\cite{yu2020reclor}, 
ARCT and ARCT\_adv~\cite{schuster2019towards} are
multiple choice reasoning tasks. 
We set the minimum number appearance of a feature
in either the training or the test set to be 5 to qualify as a cue.

\subsection{Hypothesis-only Tests}
As a comparison to our framework, we first show the hypothesis-only test results
on our 4 models and 10 datasets. In this test, we apply the models trained on
full training data (with both premise and hypothesis) of the 10 datasets, and
test their accuracies on the hypothesis-only test data (by stripping
the premises from the questions in the test set). \tabref{tab:hypoonly}
shows the results, compared with the original accuracies using the full
test data. 

\begin{table}[th]
\centering
\scriptsize
\begin{tabular}{c|c|c|c|c|c} \hline
Dataset & Majority & FT & ES & BT & RB \\ \hline
\multirow{2}{*}{SNLI} & \multirow{2}{*}{33.3} &  54.43& 87.44  &  90.56 & 91.86 \\
	& &   59.83  &    59.55  &  45.7& 45.29 \\ \hline
\multirow{2}{*}{QNLI}  & \multirow{2}{*}{50} &  67.17 & 61.60  &  86.42 & 90.37 \\
	&  &66.4   &  57.45    & 55.16 & 52.91 \\ \hline
\multirow{2}{*}{MNLI} & \multirow{2}{*}{33.3} & 47.2  & 54.63  & 83.42  & 87.21 \\
	& & 52.46&   54.57  &   36.66   &  37.84  \\ \hline
\multirow{2}{*}{ROCStory} & \multirow{2}{*}{50} & 61.73  &  62.91 &  86.85 &  91.55\\
	& &   60.24  &   59.88  & 56.44 & 74  \\ \hline
\multirow{2}{*}{COPA}   & \multirow{2}{*}{50} & 48  & 53.8  &  67.4 & 69 \\
	& &   48.4    &  51.4& 60 & 58.4\\ \hline
\multirow{2}{*}{SWAG}  & \multirow{2}{*}{25} & 27.79  &  68.95 &  77.58 &  81.89\\
	& &  27.82  &   50.62   &  53.66& 58.42 \\ \hline
\multirow{2}{*}{RACE}  & \multirow{2}{*}{25} & 29.87  &  31.35 & 29  & 29.69 \\
	& &    31.27  &  29.83    & 30.09 &  24.48\\ \hline
\multirow{2}{*}{RECLOR} & \multirow{2}{*}{25} & 32.2  & 30.96  &  45 & 54.2 \\
	& &   31.6 &  30.2    &40.2 & 32.2 \\ \hline
\multirow{2}{*}{ARCT}& \multirow{2}{*}{50} &  50.23 & 47.52  & 65.76  & 77.25 \\ 
	& &  50.23   &  49.77    & 62.83 & 65\\ \hline
\multirow{2}{*}{ARCT\_adv}& \multirow{2}{*}{50} & 50  &50 & 50.33  & 50 \\
	&  &  50  &  50    & 50 & 50\\ \hline
\end{tabular}
\caption{Hypothesis-only Tests (\%). The number on the above in each cell is the original accuracy on
full test data; on the bottom is the accuracy on hypothesis-only tests.}\label{tab:hypoonly}
\end{table}

\begin{figure}[th]
\centering
\includegraphics[width=\columnwidth]{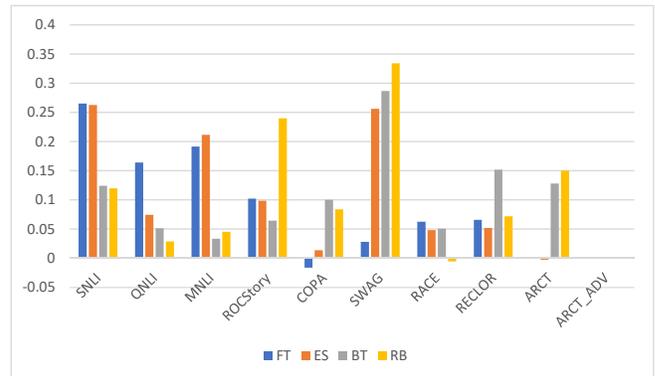}
\caption{Four models' hypothesis-only tests accuracies beyond vote by majority (hypo-only
acc $-$ majority acc)}
\label{fig:ending1}
\end{figure}

We further plot the four models hypothesis-only accuracies against voting by majority
results in \tabref{tab:hypoonly} in \figref{fig:ending1}. 
This figure depicts the ``weakness'' of the datasets to these models. 
The higher the bars, the weaker the dataset. We can see that SWAG, SNLI and MNLI
are generally easier, whereas ARCT\_ADV is a hard task (the deviation results
approach to zero).

\begin{figure}[th]
\centering
\includegraphics[width=\columnwidth]{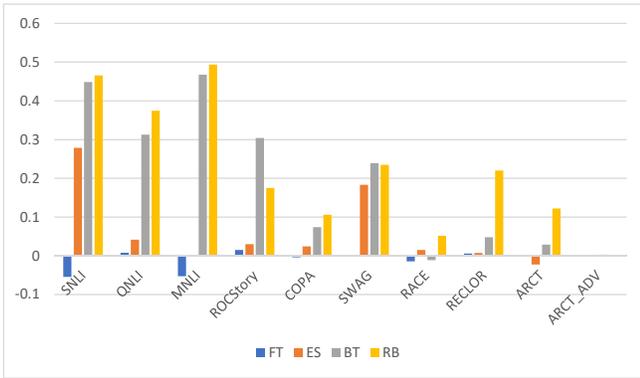}
\caption{Four models' full test accuracies above hypothesis-only tests accuracies (full acc $-$
hypo-only acc)}
\label{fig:ending2}
\end{figure}

Next, we plot the differences between the model accuracies on the full test data
and the accuracies on the hypothesis-only data in \figref{fig:ending2}. 
This experiment evaluates the robustness of the models. 
We see that the bars for FastText or ESIM are very short, while
the bars are much longer for BERT and RoBERTA. This shows that from 
the hypothesis-test point of view, BERT and RoBERTA are more robust 
against the artifacts in the datasets, because the duo do not 
perform as well when given only
the ending of the questions. 
These preliminary results serve as the basis for our findings next.

\subsection{Cues in Datasets}

We first filter the training and test data for each dataset using all the features
defined in this paper. Left half of \tabref{tab:bias} shows the top 5 cues we have discovered
as well as their cueness score for each of the 10 datasets. ARCT\_adv is an adversarial dataset
and is known to be well balanced on purpose. As a result, we only managed to find one cue
which is OVERLAP and its cueness score is very low. This is not surprising, because
OVERLAP is the only ``second-order'' feature in our list of linguistic features that
is concerned with tokens in both the premise and hypothesis, and likely escaped from
the data manipulation by the creator. 

\begin{table}[th]
\centering
\scriptsize
\begin{tabular}{c|c|c|c|c|c|c} \hline
Dataset & Top Cues & Cueness & FT  & ES & BT & RB\\ 
	&	& \%	& ($\Delta$)& ($\Delta$)& ($\Delta$)& ($\Delta$) \\ \hline
\multirow{5}{*}{SNLI} 
& ``sleeping'' & 13.95 & 30.3 &6.81 & 5.34& 4.87 \\                                                                    
& ``no'' & 13.33 & 18.09 &3.32 & 2.05& 2.6 \\
& ``because'' & 9.24 & 18.89 &4.88 & 5.61& 4.31 \\
& ``friend'' & 8.82 & 22.96 &6.66 & 3.51& 3.05 \\
& ``movie'' & 7.73 & 16.64 &0.06 & 9.47& -0.19 \\
	   \hline 
\multirow{5}{*}{QNLI} 
& ``dioxide'' & 4.52 & 9.78 &-0.06 & 4.97& 10.56 \\                                                                    
& ``denver'' & 4.26 & 13.59 &7.14 & 2.23& 3.11 \\
& ``kilometre'' & 4.24 & 4.85 &6.43 & 4.67& 2.55 \\
& ``mile'' & 3.95 & 7.16 &15.64 & -1.65& -6.65 \\
& ``newcastle'' & 3.8 & 3.44 &12.0 & 0.89& -1.23 \\
	   \hline 
\multirow{5}{*}{MNLI} 
& ``never'' & 10.4 & 29.15 &26.41 & 9.86& 10.6 \\                                                                      
& ``no'' & 8.98 & 19.49 &20.17 & 1.2& 3.32 \\
& ``nothing'' & 8.98 & 25.5 &26.84 & 5.11& 4.32 \\
& ``any'' & 6.79 & 20.4 &19.39 & 7.76& 3.74 \\
& ``anything'' & 5.73 & 18.43 &15.74 & 3.31& 1.14 \\
	   
	   \hline 
\multirow{5}{*}{ROCStory} 
& ``threw'' & 12.99 & 1.28 &4.69 & 10.88& 0.97 \\                                                                      
& ``now'' & 8.68 & -10.01 &14.51 & 1.75& 5.69 \\
& ``found'' & 8.16 & -2.31 &4.45 & 5.12& -3.13 \\
& ``won'' & 7.71 & 2.43 &0.74 & 1.05& 5.51 \\
& ``like'' & 7.3 & 4.77 &10.06 & 8.81& 1.67 \\
	   \hline 
\multirow{5}{*}{COPA} 
& ``went'' & 3.61 & -10.83 &6.46 & 7.92& 1.04 \\                                                                       
& ``got'' & 2.74 & 5.45 &-9.89 & -12.52& -10.3 \\
& ``for'' & 2.14 & 10.11 &-1.89 & 9.05& 11.58 \\
& ``with'' & 1.38 & -15.64 &-6.98 & 3.3& 13.82 \\
& TYPO & 0.84 & -12.46 &-2.33 & 3.8& -8.22 \\
	    \hline 
\multirow{5}{*}{SWAG}
& ``football'' & 7.38 & 6.13 &8.55 & 1.2& 1.55 \\
& ``anxious'' & 6.65 & 7.55 &-4.67 & -6.66& -1.67 \\
& ``concerned'' & 6.19 & 12.6 &4.58 & 8.27& -5.66 \\
& ``skull'' & 5.73 & -2.77 &0.49 & 8.43& 3.49 \\
& ``cop'' & 5.01 & 2.79 &5.3 & -0.92& -0.04 \\
\hline 

\multirow{5}{*}{RACE} 
& ``above'' & 13.74 & 8.73 &-8.43 & -0.22& -1.92 \\                                                                    
& ``b'' & 12.84 & 16.97 &-4.8 & 3.52& -3.45 \\
& ``c'' & 11.83 & 15.69 &-6.94 & 8.6& -7.6 \\
& ``probably'' & 6.77 & 9.91 &-0.06 & -3.8& 2.86 \\
& ``may'' & 4.2 & 7.75 &-3.45 & -6.67& -1.8 \\
	   
	   \hline 
\multirow{5}{*}{RECLOR} 
& ``over'' & 2.07 & 1.76 &-2.94 & -1.35& -4.12 \\                                                                      
& ``result'' & 1.97 & -3.29 &-2.69 & -1.78& -3.7 \\
& ``explanation'' & 1.81 & -6.33 &-1.73 & -2.76& -7.24 \\
& ``proportion'' & 1.68 & -5.64 &-4.69 & 2.37& -2.16 \\
& ``produce'' & 1.4 & 4.54 &-2.98 & -14.36& -3.7 \\
	   \hline 
\multirow{5}{*}{ARCT} 
& ``not'' & 3.74 & -2.54 &7.45 & -0.97& -11.96 \\                                                                      
& NEGATION & 2.85 & 3.49 &10.04 & 6.28& -8.23 \\
& ``n't'' & 2.52 & 10.3 &5.89 & 9.49& 4.84 \\
& ``always'' & 2.25 & -4.66 &38.21 & -4.35& -8.26 \\
& ``doe'' & 2.06 & -0.73 &-3.69 & -1.15& -7.22 \\
	   \hline 
ARCT\_adv& OVERLAP & 1.96e-10 & 1.65 &-0.25 & 2.73& 0.57 \\\hline
\multicolumn{3}{c|}{$\sum(|.|)$ (Model weakness)} 	& 469.8 & 361.4 & 227.7 & 216.2 \\
\hline 
\end{tabular}
\caption{Datasets, their top 5 cues and 4 models biases $\Delta$ on them.}\label{tab:bias}
\end{table}

In most of the datasets, the top 5 cues discovered are word features,
but besides OVERLAP, we do see NEGATION and TYPO showing up
in the lists. In fact, SENTIMENT and NER features would have shown up
if we expanded the list to top 10. It is also interesting to see several features previously
reported to be biased by other works, such as ``not'' and NEGATION in
ARCT, ``no'' in MNLI and SNLI, and ``like'' in ROCStory.  Especially in
MNLI, all the five cues discovered are related to negatively toned words,
suggesting significant human artifacts in this datasets.
 
In the results, we also identify that some of the word cues are indicative of
certain syntactic/semantic/sentiment patterns in the questions. For example, ``because'' in SNLI
indicates a causal-effect structure; ``like'' in ROCStory indicates positive sentiment;
``probably'' and ``may'' in RACE indicate uncertainty, etc. 

If we sum up the top 5 cueness scores for each dataset, we find that
overall SNLI, RACE, ROCStory and MNLI carry more cueness than others,
while ARCT\_adv has the smallest cueness. This result is
generally consistent with our findings in the hypothesis-only test (see \figref{fig:ending1})
earlier,
though now we can pinpoint what causes the weaknesses in these four
datasets. The only exception is SWAG, which didn't emerge as a very
weak dataset in the experiment here, but was found very biased in
the hypothesis-only test. The reason is SWAG is the biggest dataset
among the ten and we discovered more cues in it than
the other 9 combined. Therefore, the top 5 cues are insufficient
to represent the full scale of its weakness, which is showed by
the hypothesis test as it gives the full picture.

\begin{figure*}[th]
\centering
\begin{subfigure}[b]{0.3\textwidth}
\centering
\includegraphics[width=\columnwidth]{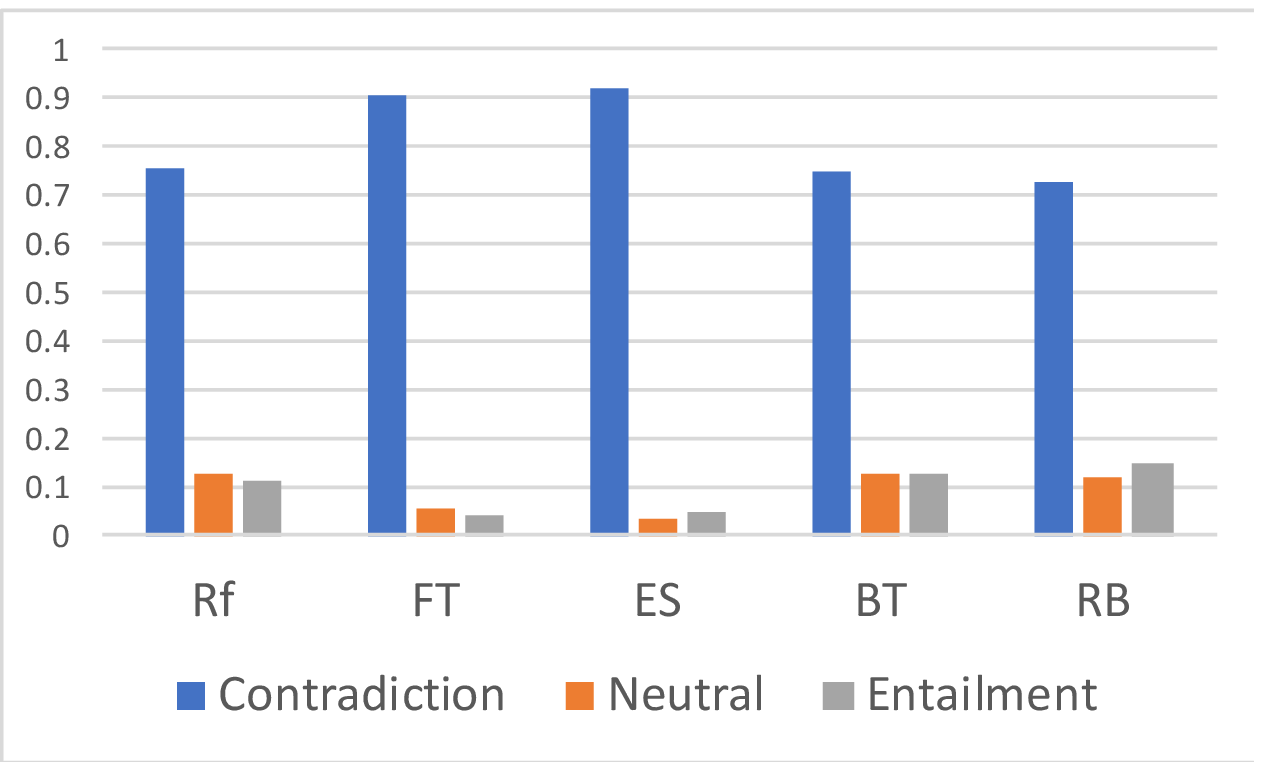}
\caption{Cue ``no'' in MNLI}
\label{fig:cue_no}
\end{subfigure}
\hfill
\begin{subfigure}[b]{0.3\textwidth}
\centering
\includegraphics[width=\columnwidth]{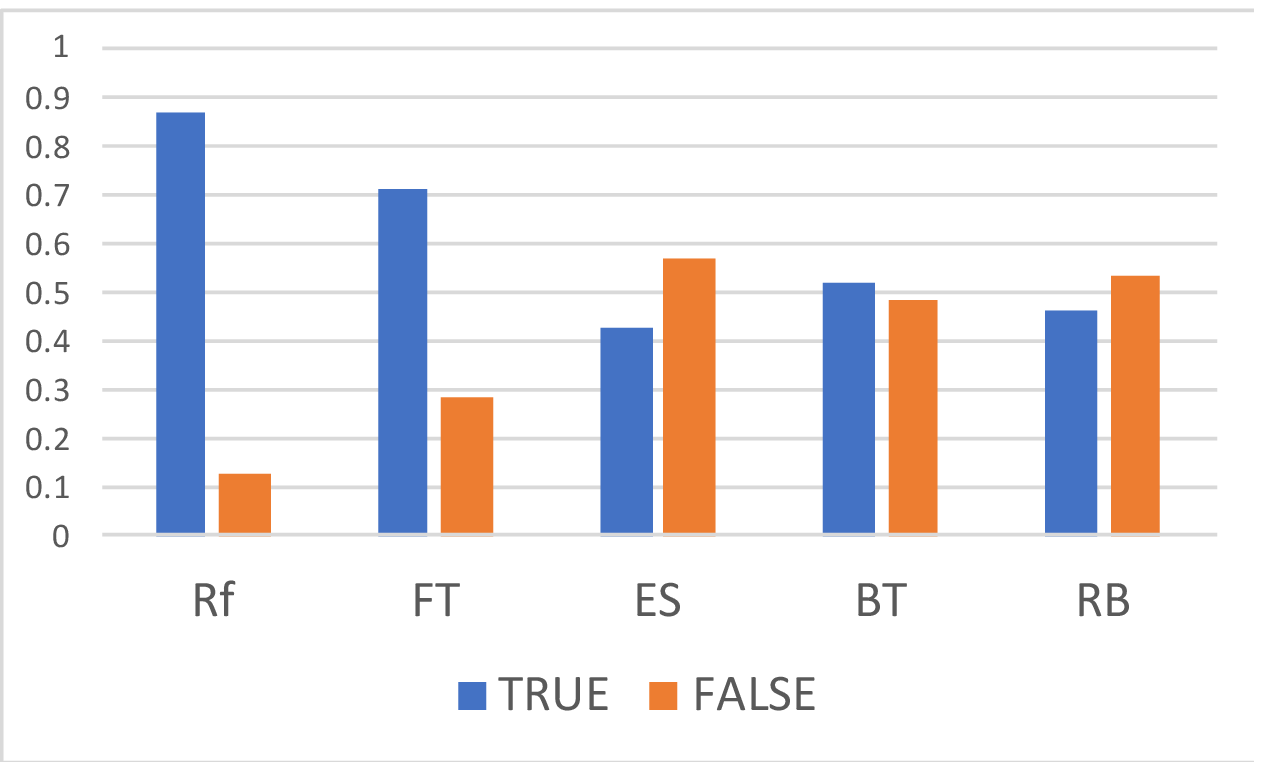}
\caption{Cue ``above'' in ARCT}
\label{fig:cue_above}
\end{subfigure}
\hfill
\begin{subfigure}[b]{0.3\textwidth}
\centering
\includegraphics[width=\columnwidth]{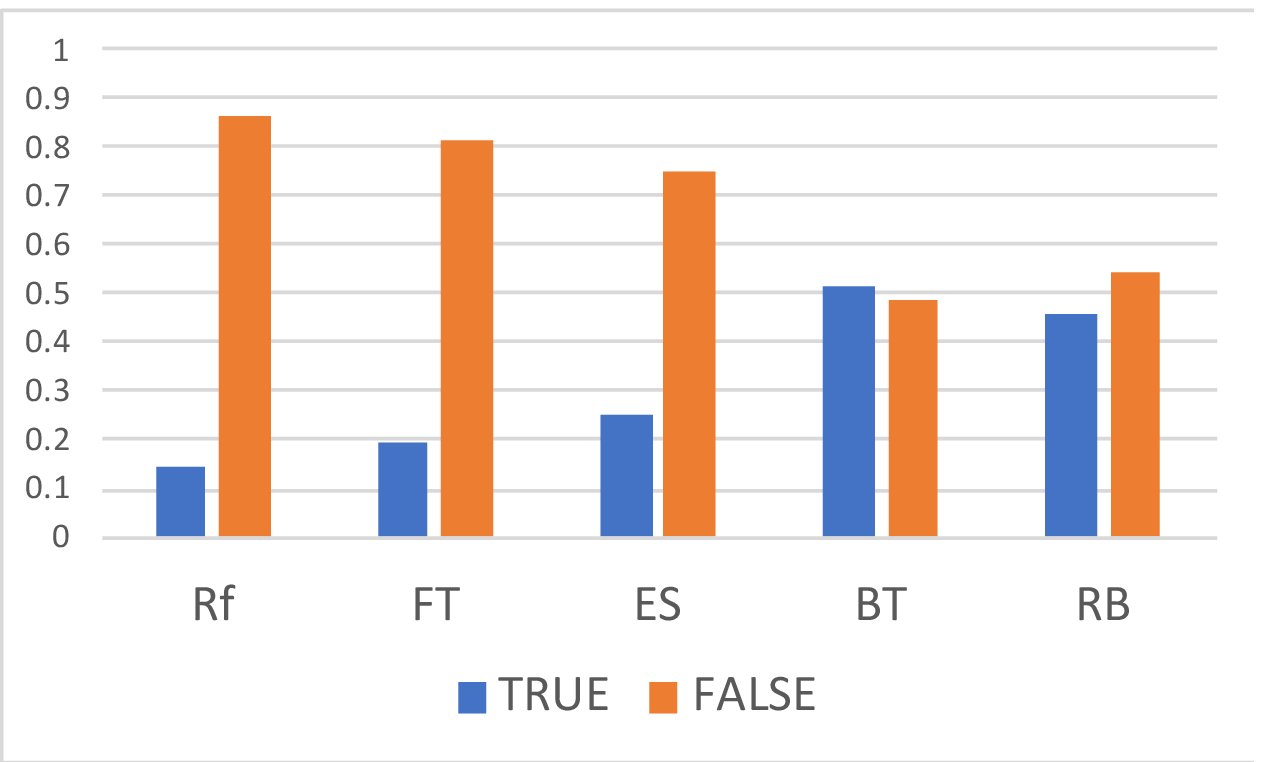}
\caption{Cue ``threw'' in ROCStory}
\label{fig:cue_threw}
\end{subfigure}
\caption{Three test examples for distribution comparison with 4 different models}
\label{fig:cue_result}
\end{figure*}

\subsection{Biases in Models}

For each feature and a dataset, 
we train four models on its original training set,
and test the models on its full test set, feature-filtered test set, and neutral test set.   
We first show the result of ``accuracy tests'' in \tabref{tab:bias}.
If $\Delta$ is positive for a model on a feature, it means that
the model exploits the existence of this feature. Conversely,
the model exploits the non-existence of the this feature.
A model is more robust against biases in the data, if $\Delta$ is
close to zero.
Therefore, the bottom of \tabref{tab:bias} shows that,
across all 10 datasets, by the sum of the absolute values of $\Delta$, 
RoBERTA $<$ BERT $<$ ESIM $<$ FastText. This again is consistent with
our hypothesis-only test earlier and 
the community's common perception of these popular models.
However, if we delve into individual datasets and features, 
situation can be a bit murkier.

For example, it seems that FastText tends to pick up more individual word cues
than semantic cues, but more complex models such as BERT
and RoBERTA appear to be more sensitive to structural features such as NEGATION
and SENTIMENT, which are actually classes of words. 
This can be well explained by the way FastText is designed to
model words more than syntactic or semantic structures.

The fact that FastText is strongly negatively correlated with TYPO is
also interesting. We speculate that FastText might have been
trained with a more orthodoxical vocabulary and thus less
tolerant to typos in text. 

We further invesigate the models using the
``distribution test''. 
We show three insteresting findings in \figref{fig:cue_result}. 
We observe that all models on cue ``no'' in MNLI 
achieve positive $\Delta$ in \tabref{tab:bias}, and fastText in particular. 
Consistent with the ``Accuracy test'', we find the predicting label distribution 
skewness is amplified in \figref{fig:cue_no} for fastText and ESIM -  
with ``no'' in sight, they prefer to predict ``Contradiction'' even more
than the ground truth in training data.
On the contrary, BERT and RoBERTA are only moderate in following
the training data. 

If cue ``no'' is very good at tricking the models,
cue ``above'' is not as successful. 
\figref{fig:cue_above} shows that 
the distribution of predicting result for ESIM in ARCT 
is completely opposite to the training data. 
This explains while $\Delta=-8.43$ in \tabref{tab:bias} and
demonstrates that models may not take advantage of the cue even though it is
right there in the data.
Similarly, the ``flatness'' in BERT and RoBERTA 
can also explain their low $\Delta$ values in \tabref{tab:bias}. 

The example of cue ``threw'' presents an outlier for BERT,
because the distribution test result is inconsistent between the accuracy test: 
The accuracy deviation is very high for BERT, but its prediction distribution is
flat. We haven't seen a lot of such contradictory cases so far. 
But when it happens, as it is here, we give BERT the benefit of the doubt 
that it might not have exploited the cue ``threw''. 


%

\section{Related Work}
\label{sec:related}

Our work is related to and, to some extent, comprises of 
elements in three research directions: spurious features analysis, 
bias calculation and dataset filtering. 
 
\textbf{Spurious features analysis} has been increasingly studied recently. 
Much work~\cite{sharma2018tackling,srinivasan2018simple,zellers2018swag} 
has observed that some NLP models can surprisingly 
get good results on natural language understanding questions in MCQ form without 
even looking at the stems of the questions. Such tests are called
``hypothesis-only'' tests in some works.
Further, some research~\cite{sanchez2018behavior} discovered that these models 
suffer from an insensitivity to certain small but semantically significant alterations
in the hypotheses, leading to speculations that the hypothesis-only performance
is due to simple statistical correlations between words in the hypothesis 
and the labels. 
Spurious features can be classified into
lexicalized and unlexicalized~\cite{bowman2015large}:
lexicalized features mainly contain indicators of n-gram tokens and cross-ngram tokens, 
while unlexicalized features involve word overlap, sentence length and BLUE score between 
the premise and the hypothesis. ~\citealp{naik2018stress} refined the 
lexicalized classification to Negation, Numerical Reasoning, 
Spelling Error. ~\citealp{mccoy2019right} refined the word overlap 
features to Lexical overlap, Subsequence and Constituent 
which also considers the syntactical structure overlap. ~\citealp{sanchez2018behavior} 
provided unseen tokens an extra lexicalized feature. 

\textbf{Bias calculation} is concerned with methods to quantify the severity of the cues. 
Some work~\cite{clark2019don,he2019unlearn,yaghoobzadeh2019robust} 
attempted to encode the cue feature implicitly by 
hypothesis-only training or by extracting features associated with a certain label 
from the embeddings. 
Other methods compute the bias by statistical metrics. 
For example, \citealp{yu2020reclor} used the probability of seeing a word 
conditioned on a specific label to rank the words by their biasness. 
LMI~\cite{schuster2019towards} was also used to evaluate cues and 
re-weight in some models. 
However, these works did not give the reason to use these metrics, one way or 
the other.
Separately, \citealp{Marco2020acl} gave a test data augmentation method, 
without assessing the degree of bias in those datasets.
 
\textbf{Dataset filtering} is one way of achieving
higher quality in datasets by reducing artifacts. 
In fact, datasets such as SWAG and RECLOR evaluated in this paper
were produced using variants of this filter approach which 
iteratively perturb the data instances until a target 
model can no longer fit the resulting dataset well. 
Some methods~\cite{yaghoobzadeh2019robust}, instead of preprocessing
the data by removing biases, leave out samples with biases 
in the middle of training according to decision made between 
epoch to epoch. \citealp{bras2020adversarial} investigated 
model-based reduction of dataset cues and designed an algorithm 
using iterative training. Any model can be used in 
this framework. Although such an approach is more 
general and more efficient than human annotating, 
it heavily depends on the models. Unfortunately, different models
may catch different cues. Thus, such methods may not be complete.

\section{Conclusion}
We develop a light-weight 
framework that evaluates the potential biases and cues in NLR multiple choice 
datasets and further shed light on
the exploration of models at least from the perspective of the statistical cues. 
We experimented on a large range of datasets covering different tasks and 
conclude that the new evaluation framework is effective in discovering
bias problems in both the datasets and some popular models.

\bibliographystyle{named}
\bibliography{aaai21}

\end{document}